\documentclass[10pt,twocolumn,letterpaper]{article}
\usepackage[accsupp]{axessibility} 
\usepackage{iccv}
\usepackage{times}
\usepackage{epsfig}
\usepackage{graphicx}
\usepackage{amsmath}
\usepackage{amssymb}
\usepackage{commath}
\usepackage{multirow}
\usepackage{amsfonts,amsthm}
\usepackage{subcaption}
\usepackage{comment}
\usepackage[ruled,vlined]{algorithm2e}
\usepackage{xcolor}
\usepackage{bm, epstopdf, url, pifont, overpic, cases}

\usepackage[pagebackref=true,breaklinks=true,letterpaper=true,colorlinks,bookmarks=false]{hyperref}
\usepackage[nocompress]{cite}

\iccvfinalcopy 

\def\iccvPaperID{} 

\ificcvfinal\fi

\begin{document}

\title{Restore Anything Pipeline: Segment Anything Meets Image Restoration

}

\author{Jiaxi Jiang\qquad\quad Christian Holz\\
Department of Computer Science, ETH Zurich, Switzerland\\
{\tt\small \{jiaxi.jiang, christian.holz\}@inf.ethz.ch}\\
\url{https://github.com/eth-siplab/RAP}
}

\maketitle

\ificcvfinal\thispagestyle{empty}\fi

\begin{abstract}
Recent image restoration methods have produced significant advancements using deep learning. 
However, existing methods tend to treat the whole image as a single entity, failing to account for the distinct objects in the image that exhibit individual texture properties.
Existing methods also typically generate a single result, which may not suit the preferences of different users.
In this paper, we introduce the \emph{Restore Anything Pipeline} (RAP), a novel interactive and per-object level image restoration approach that incorporates a controllable model to generate different results that users may choose from.
RAP incorporates image segmentation through the recent Segment Anything Model (SAM) into a controllable image restoration model to create a user-friendly pipeline for several image restoration tasks.
We demonstrate the versatility of RAP by applying it to three common image restoration tasks: image deblurring, image denoising, and JPEG artifact removal.
Our experiments show that RAP produces superior visual results compared to state-of-the-art methods. 
RAP represents a promising direction for image restoration, providing users with greater control, and enabling image restoration at an object level.
\end{abstract}

\section{Introduction}

\begin{figure}[t!]
    \centering

\includegraphics[trim=0cm 0cm 0cm 0cm,clip=true,width=0.32\linewidth]{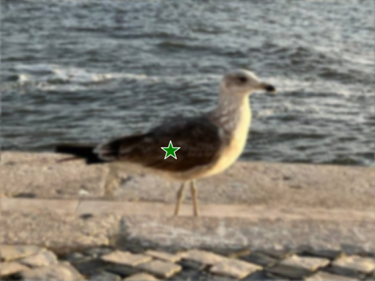}
\includegraphics[trim=0cm 0cm 0cm 0cm,clip=true,width=0.32\linewidth]{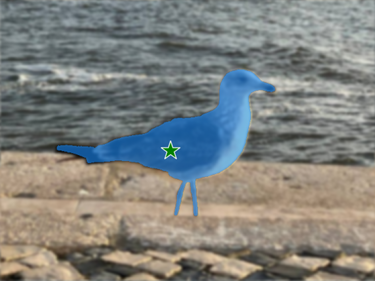}
\includegraphics[trim=0cm 0cm 0cm 0cm,clip=true,width=0.32\linewidth]{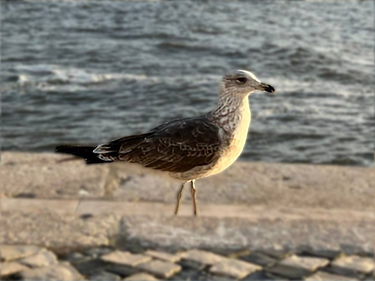}

\includegraphics[trim=0cm 0cm 0cm 0cm,clip=true,width=0.32\linewidth]{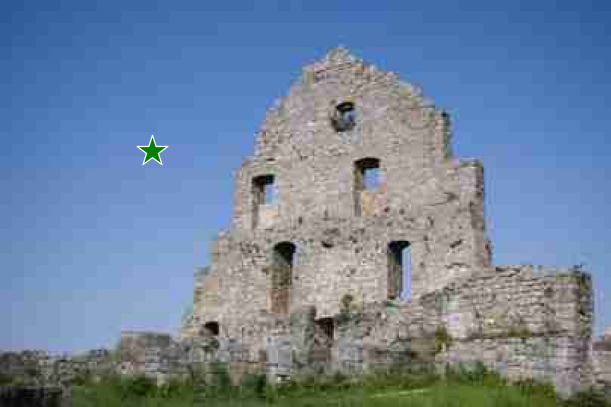}
\includegraphics[trim=0cm 0cm 0cm 0cm,clip=true,width=0.32\linewidth]{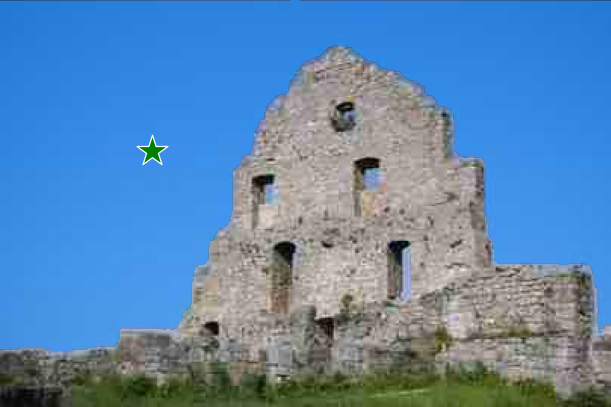}
\includegraphics[trim=0cm 0cm 0cm 0cm,clip=true,width=0.32\linewidth]{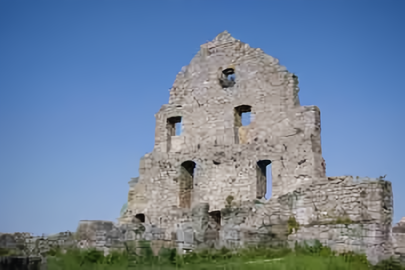}

\includegraphics[trim=0cm 0cm 0cm 0cm,clip=true,width=0.32\linewidth]{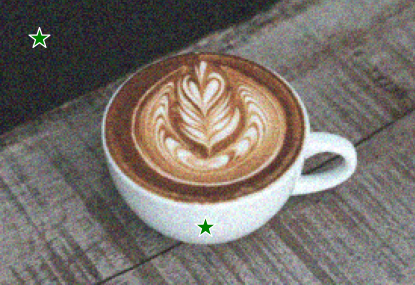}
\includegraphics[trim=0cm 0cm 0cm 0cm,clip=true,width=0.32\linewidth]{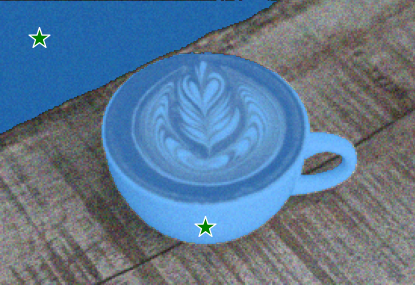}
\includegraphics[trim=0cm 0cm 0cm 0cm,clip=true,width=0.32\linewidth]{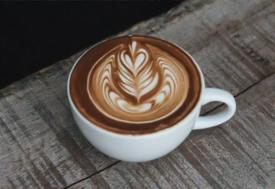}

    \caption{Our proposed \emph{Restore Anything Pipeline} (RAP) is capable of deblurring, denoising, and removing JPEG artifacts.
    	By selecting objects, RAP restores and enhances images at an per-object level, removing noise while preserving important texture details for high-quality results.
    	RAP's per-object sensitivity is particularly important due to each object's individual texture and thus noise characteristics.
    	Please zoom in for visual details.}
    \label{fig:rap_example}
\end{figure}
Image restoration has long been a significant challenge due to its practical importance in various low-level vision applications.
The primary objective of image restoration is to recover the original high-quality image from its degraded observation. Image restoration is an ill-posed inverse problem since there can be many possible candidates for the same low-quality input. By specifying different degradation operations, we can address specific image restoration tasks like image deblurring~\cite{kupyn2018deblurgan, ren2020neural, zhang2020deblurring}, image denoising~\cite{zhang2017beyond, guo2019toward,wang2023lg}, JPEG artifact removal~\cite{jiang2021towards, ehrlich2020quantization, fu2019jpeg}, and image super-resolution~\cite{zhang2018residual,liang2021swinir,zhang2021designing}. Besides, image restoration techniques can be widely applied into other tasks such as remote sensing~\cite{jiang2019edge,dong2020remote}, capacitive sensing~\cite{streli2021capcontact, mayer2021super}, and biomedical imaging~\cite{schnitzbauer2017super, chen2020unsupervised}.

Despite significant advancements in image restoration, several challenging problems persist.
First, current methods typically treat the entire image as a single entity, disregarding the presence of different objects with distinct texture properties within an image.
For instance, an image may consist of objects with both high-frequency signal components and smooth objects with low-frequency regions.
While noise tends to manifest as a high-frequency signal, making it easily observable in low-frequency regions, it becomes challenging to distinguish from high-frequency texture details.
Conversely, a strong denoising algorithm may perform well on objects with low-frequency regions, such as cartoon images,
but may result in over-smoothed outcomes on high-frequency regions due to the removal of texture details.
Thus, striking a balance between noise removal and detail preservation presents a significant challenge.
Second, most existing methods are limited to providing a single deterministic result.
However, in the real world, evaluating image quality is often subjective, as individual users may have varying preferences.
Therefore, it would be advantageous to have a single model capable of generating diverse results, allowing users to select the output that best suits their requirements. 

To address the first problem, we introduce a novel pipeline that integrates Segment Anything Model~\cite{kirillov2023segment} into the restoration process. 
Our pipeline enables object-level controllable image restoration, allowing for independent treatment of objects with diverse texture details.
Therefore, our approach avoids generating a generic or ``mean average'' result across all parts and, thus, improves the overall quality of the restoration process.
To address the second problem, we utilize a flexible blind image restoration framework~\cite{jiang2021towards} that offers the unique advantage of accommodating both blind and non-blind models. Users can effortlessly obtain fully automatic restored results without knowing the degradation parameter, but also have the flexibility to adjust the predicted degradation parameters, granting precise control over the output outcomes.
We demonstrate our approach with a primary focus on three fundamental image restoration tasks: image deblurring, image denoising, and JPEG artifact removal.

Collectively, we make the following main contributions: 

(1)~We propose the \emph{Restore Anything Pipeline} (RAP), which incorporates the recent Segment Anything Model to segment individual objects before performing semantic image restoration on a per-object level.
To the best of our knowledge, this is the first attempt to use image segmentation to help image restoration in an interactive way.

(2)~For interactive image restoration, we extend the flexible blind JPEG artifact removal model FBCNN to perform general image restoration tasks, including image denoising and deblurring.
In our approach, image restoration can be both automatic or controlled by users.

(3)~We demonstrate the effectiveness of RAP on real images with different degradation settings, representing a promising direction for image restoration.
\section{Related Work}

\paragraph{Interactive Image Processing.}

Interactive image processing refers to a process where users can actively participate in image processing, allowing for greater flexibility and customization of the results. This feature is commonly found in commercial image editing software for performing basic enhancement operations. It is also embraced by advanced intelligent image editing techniques~\cite{pan2023_DragGAN, lee2020maskgan, ling2021editgan, kawar2023imagic}.
In image restoration, FFDNet~\cite{zhang2018ffdnet} takes a tunable noise level map as the input to remove noise with different levels. SRMD~\cite{zhang2018learning} takes blur kernel and noise level as input for super-resolution. Wang~\etal proposed a controllable framework for interactive image restoration~\cite{wang2019cfsnet}. In~\cite{he2020interactive, cai2021toward},  multi-dimension modulation frameworks were proposed for controllable image restoration. However, these methods usually assume that the controllable variable is provided, but such information is almost unknown in real applications. To solve this problem, FBCNN~\cite{jiang2021towards} predicts the degradation parameter of the input image and use it to guide image restoration process. The predicted quality factor can also be tuned manually to control the level of restoration. However, it only targeted on JPEG images so it remains unclear if such a flexible blind framework also works for other tasks like image denoising and image deblurring. Besides, the existing controllable models are always applied to the whole image, without the consideration of different semantic difference.

\paragraph{Image Restoration and Segmentation}
Image segmentation and image restoration are two fields in computer vision and they are usually studied independently. As a low-level tasks, image restoration can be used as a preprocessing step or integrated into a pipeline for downstreaming tasks including image segmentation~\cite{dai2016image, wang2020dual}, image classification~\cite{pei2019effects,wang2020deep}, object detection~\cite{haris2021task, sun2022rethinking}, motion estimation~\cite{shariati2020towards,rozumnyi2022motion}. 
However, there are only a few works on segmentation for image restoration. For example, segmentation maps are used as additional inputs in~\cite{chen2021progressive, wang2018recovering} to guide image restoration. However, these method assume perfect segmentation maps are available and can only provide deterministic results. Instead, we integrate image segmentation into our pipeline and support interactive image restoration.

\paragraph{Segment Anything Model.}
Segment Anything Model (SAM)~\cite{kirillov2023segment} is a recently proposed strong segmentation foundation model that can produce high-quality segmentation masks by a variety of input prompts such as points, a rough box or mask, free-form text, or any information indicating what to segment in an image. Subsequently, multiple follow-up studies have emerged, extending the application of the SAM model to various tasks. These include medical image analysis~\cite{ma2023segment}, object tracking~\cite{cheng2023segment}, image editing~\cite{xie2023edit}, and image inpainting~\cite{yu2023inpaint}. In this work, we show this advance of image segmentation will benefit restoration.

\begin{figure*}
    \centering
    \includegraphics[width=1.0\linewidth]{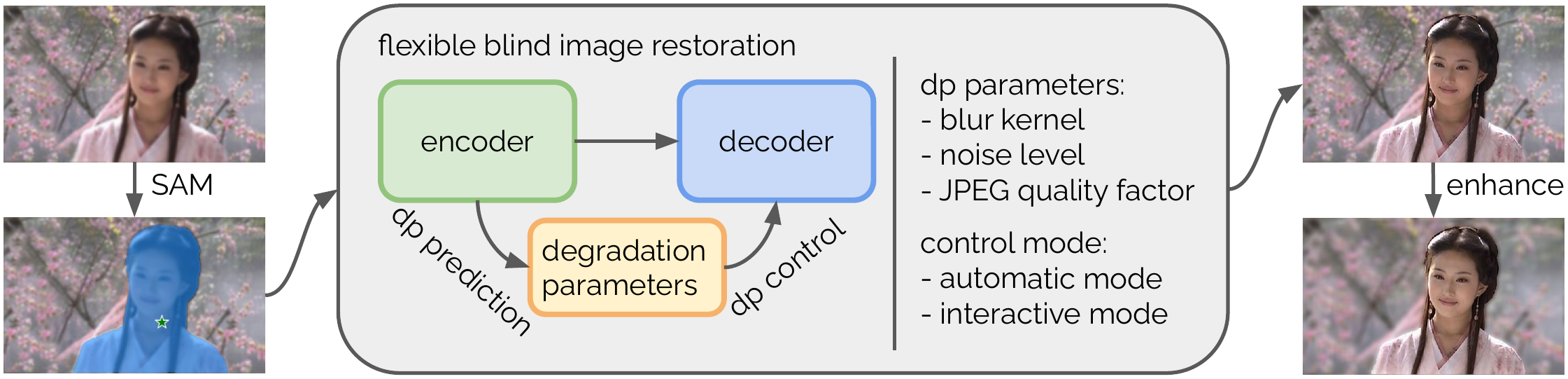}
    \caption{\emph{Restore Anything Pipeline} (RAP).
By clicking on the desired objects, RAP utilizes the SAM model to do segmentation. The segmented objects are then restored using a flexible blind image restoration framework. This framework leverages the model's prediction of the degradation parameter in the low-quality input image to guide the restoration process. Additionally, users have the option to manually adjust the predicted degradation parameters, allowing for customization and the generation of different results according to their preferences.}
    \label{fig:rap}
\end{figure*}

\section{Method}
The architecture of \emph{Restore Anything Pipeline} (RAP) is depicted in Figure~\ref{fig:rap}. It comprises three steps: image segmentation, image restoration, and image enhancement. By selecting the desired objects, RAP employs the SAM~\cite{kirillov2023segment} model to perform segmentation. The segmented objects are subsequently restored using a flexible blind image restoration framework. This framework takes advantage of the model's prediction of the degradation parameter in the low-quality input image to guide the restoration process. Moreover, users have the flexibility to manually fine-tune the predicted degradation parameters, enabling customization and generating diverse results based on their preferences. After restoration, users can also post-enhance different objects also segmented via SAM to further refine and improve the overall image quality.

\subsection{Interactive Image Segmentation}
We employ the recently introduced Segment Anything Model (SAM)~\cite{kirillov2023segment} to segment our desired object. SAM has three components: (a) an image encoder that is based on a MAE~\cite{he2022masked} pre-trained Vision Transformer~\cite{dosovitskiy2021an}, (b) a flexible prompt encoder that is divided into sparse prompts (i.e., points, boxes, text)) and dense prompts (i.e., masks), and (c) a fast mask decoder that is based on Transformer decoder and can efficiently map the image embedding, prompt embeddings, and an output token to a mask.

We have discovered that the Segment Anything Model (SAM) demonstrates remarkable robustness even when faced with corrupted images, thereby ensuring the seamless functioning of our pipeline. By providing input prompts such as clicking on the objects that are to be restored, SAM can segment them accurately. After segmentation, objects with different texture properties or degradation types can be handled with different restoration levels independently.

\begin{figure*}[]
    \centering
    \includegraphics[width=\linewidth]{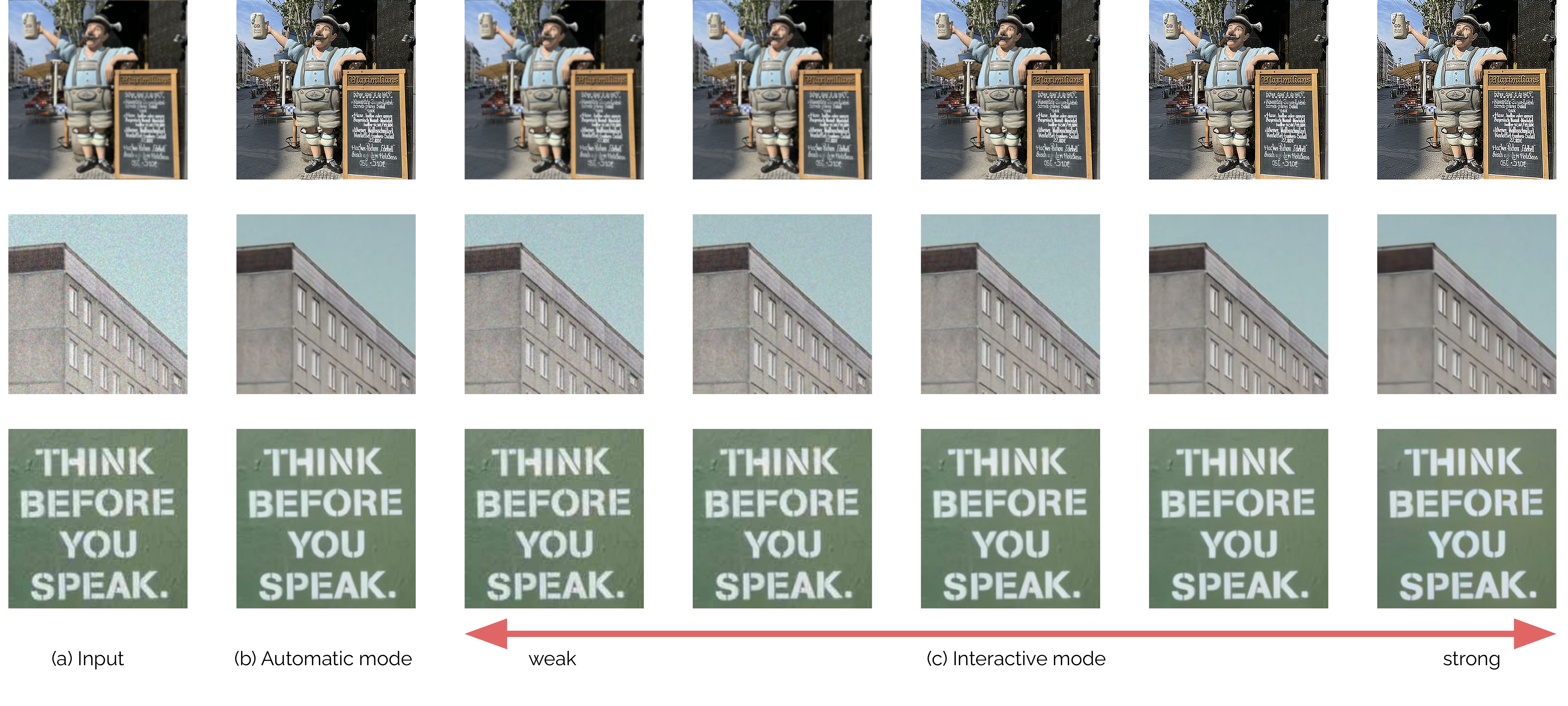}
    \caption{RAP is built up on a flexible blind image restoration framework which is capable of automatically providing restored results given a corrupted image. Additionally, users can interactively adjust the predicted parameters, enabling them to customize the level of restoration. This interactive feature enables users to select the desired results that best suit their needs. We provide examples for deblurring, denoising, and JPEG artifact removal. Please zoom in for visual details.}
    \label{fig:flexible_control}
\end{figure*}

\subsection{Object-Level Interactive Image Restoration}

To support automatic and interactive image restoration at the same time, we employ a similar framework as FBCNN~\cite{jiang2021towards}. When presented with a corrupted object, the encoder extracts deep image features and also predicts the corresponding degradation parameter.
This predicted parameter serves as a guide for the decoder, which then performs image restoration accordingly.
Importantly, users have the flexibility to manually adjust the predicted degradation parameter, enabling them to control the model and generate outputs with varying degrees of restoration, thereby accommodating individual preferences.

\paragraph{JPEG Artifact Removal}
For JPEG image artifact removal, we apply the default FBCNN model as it was originally designed for this task. The predicted degradation parameters are quality factors ranging from 0 to 100, where larger values denote higher quality.

\paragraph{Denoising}
For simplicity, we consider the commonly used Gaussian noise. To adapt FBCNN to work for image denoising, we use the standard deviation of noise distribution as predicted degradation parameter, and use this noise level to guide image restoration. We find this standard deviation can be predicted accurately and it is also controllable to provide different levels of restoration.

\paragraph{Deblurring}
We consider isotropic Gaussian kernel in our experiments. Different from image denoising and JPEG image deblocking, we find it challenging to use single parameter like the width of the blur kernel to represent the degradation information because it is difficult to predict and flexibly control the image restoration process.
Inspired by SRMD~\cite{zhang2018learning}, which concatenates a vectorized blur kernel and a blurred image as input to enable image restoration for different degradations, we use the vectorized blur kernel as the degradation parameter and use it to guide deblurring.

\subsection{Object-Level Image Enhancement}
After restoration, users can perform image enhancement also at a per-object level, such as adjusting the brightness, contrast, smoothness, etc, so that the restored image can be further polished. Although most commercial image editing software already offers similar interactive features for basic image enhancement, these operations typically operate on either the entire image or on user-drawn masked areas.

\captionsetup[sub]{font=scriptsize,labelfont={bf,sf}}
\captionsetup[subfigure]{skip=2pt} 
\begin{figure*}
\centering
\begin{subfigure}{.24\linewidth}
    \centering
    \includegraphics[trim=0cm 0cm 0cm 0cm,clip=true,width=\linewidth]{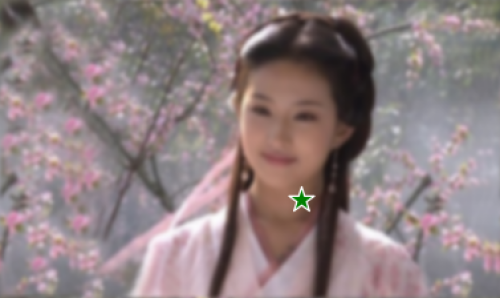}
    \caption*{Input}
\end{subfigure}
\begin{subfigure}{.24\linewidth}
    \centering
    \includegraphics[trim=0cm 0cm 0cm 0cm,clip=true,width=\linewidth]{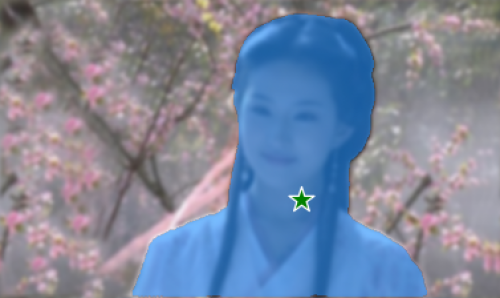}
    \caption*{SAM}
\end{subfigure}
\begin{subfigure}{.24\linewidth}
    \centering
    \includegraphics[trim=0cm 0cm 0cm 0cm,clip=true,width=\linewidth]{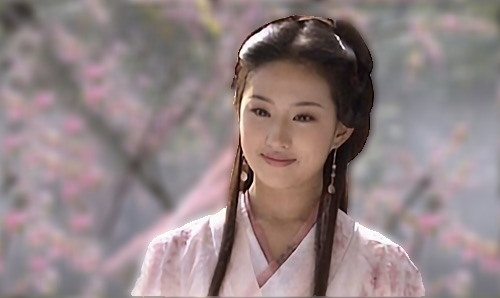}
    \caption*{RAP (Ours)}
\end{subfigure}
\begin{subfigure}{.24\linewidth}
    \centering
    \includegraphics[trim=0cm 0cm 0cm 0cm,clip=true,width=\linewidth]{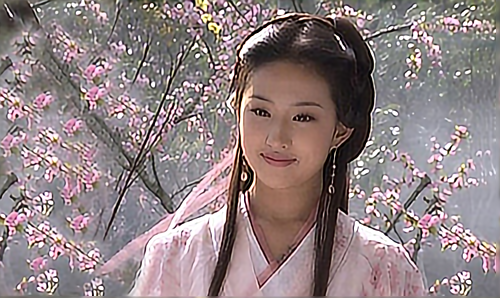}
    \caption*{FBCNN}
\end{subfigure} 

\begin{subfigure}{.24\linewidth}
    \centering
    \includegraphics[trim=0cm 0cm 0cm 0cm,clip=true,width=\linewidth]{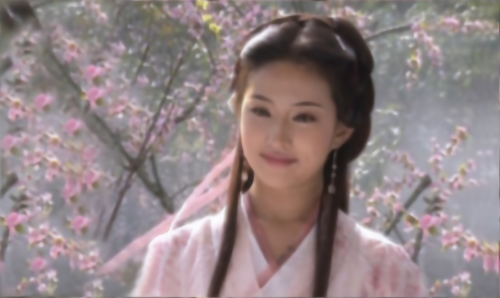}
    \caption*{Restormer}
\end{subfigure}
\begin{subfigure}{.24\linewidth}
    \centering
    \includegraphics[trim=0cm 0cm 0cm 0cm,clip=true,width=\linewidth]{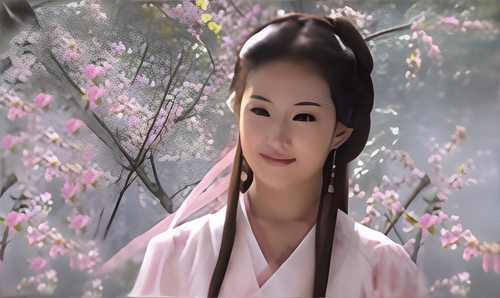}
    \caption*{Real-ESRGAN}
\end{subfigure}
\begin{subfigure}{.24\linewidth}
    \centering
    \includegraphics[trim=0cm 0cm 0cm 0cm,clip=true,width=\linewidth]{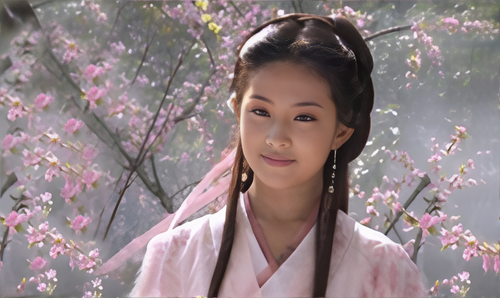}
    \caption*{GFPGAN}
\end{subfigure}
\begin{subfigure}{.24\linewidth}
    \centering
    \includegraphics[trim=0cm 0cm 0cm 0cm,clip=true,width=\linewidth]{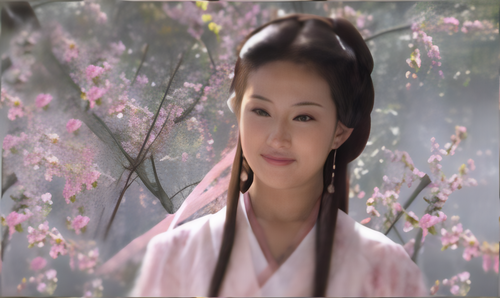}
    \caption*{BSRGAN}
\end{subfigure} 
\vspace{-5px}
\caption*{\footnotesize (a) Image deblurring}
\vspace{5px}
\begin{subfigure}{.24\linewidth}
    \centering
    \includegraphics[trim=0cm 0cm 0cm 0cm,clip=true,width=\linewidth]{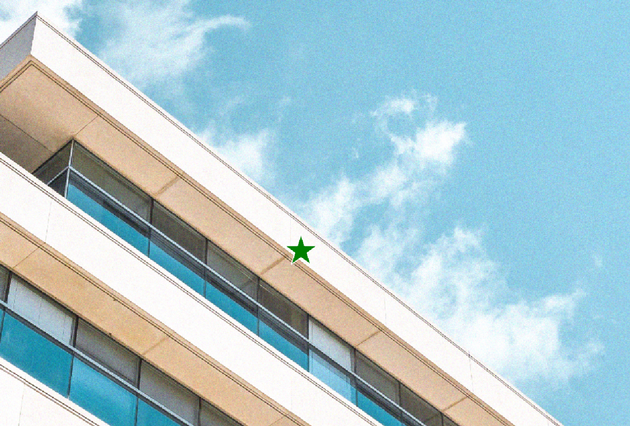}
    \caption*{Input}
\end{subfigure}
\begin{subfigure}{.24\linewidth}
    \centering
    \includegraphics[trim=0cm 0cm 0cm 0cm,clip=true,width=\linewidth]{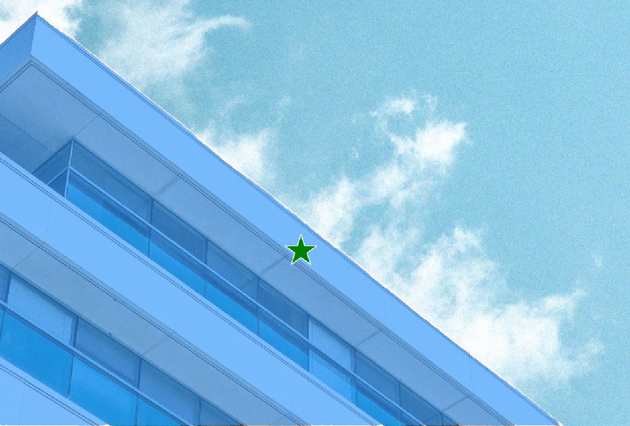}
    \caption*{SAM}
\end{subfigure}
\begin{subfigure}{.24\linewidth}
    \centering
    \includegraphics[trim=0cm 0cm 0cm 0cm,clip=true,width=\linewidth]{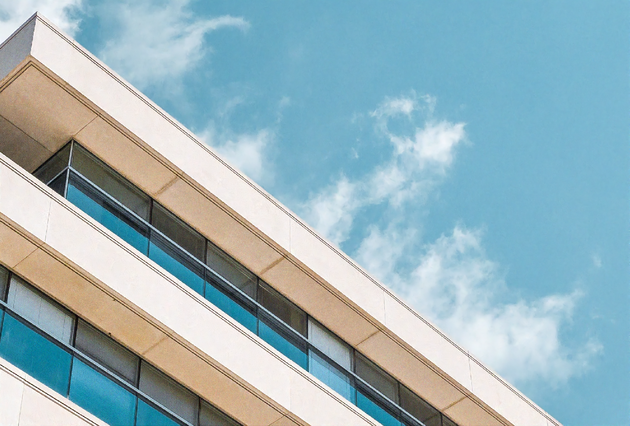}
    \caption*{RAP (Ours)}
\end{subfigure}
\begin{subfigure}{.24\linewidth}
    \centering
    \includegraphics[trim=0cm 0cm 0cm 0cm,clip=true,width=\linewidth]{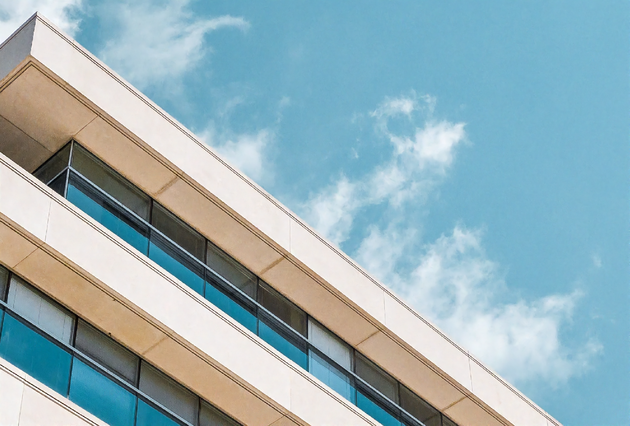}
    \caption*{FBCNN}
\end{subfigure} 

\begin{subfigure}{.24\linewidth}
    \centering
    \includegraphics[trim=0cm 0cm 0cm 0cm,clip=true,width=\linewidth]{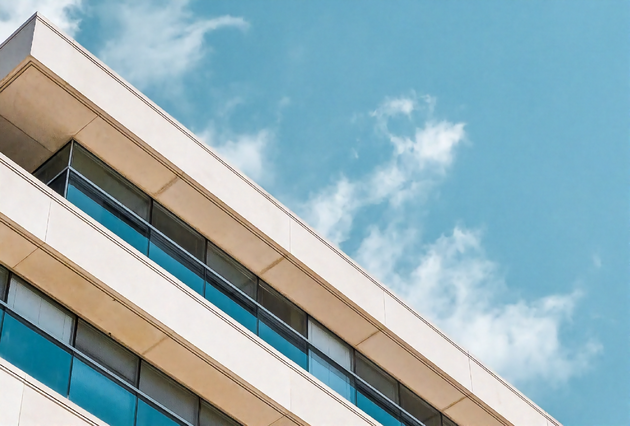}
    \caption*{FBCNN-Strong}
\end{subfigure}
\begin{subfigure}{.24\linewidth}
    \centering
    \includegraphics[trim=0cm 0cm 0cm 0cm,clip=true,width=\linewidth]{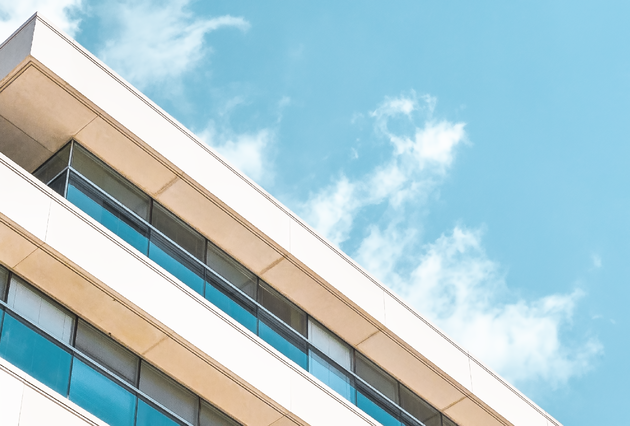}
    \caption*{Restormer}
\end{subfigure}
\begin{subfigure}{.24\linewidth}
    \centering
    \includegraphics[trim=0cm 0cm 0cm 0cm,clip=true,width=\linewidth]{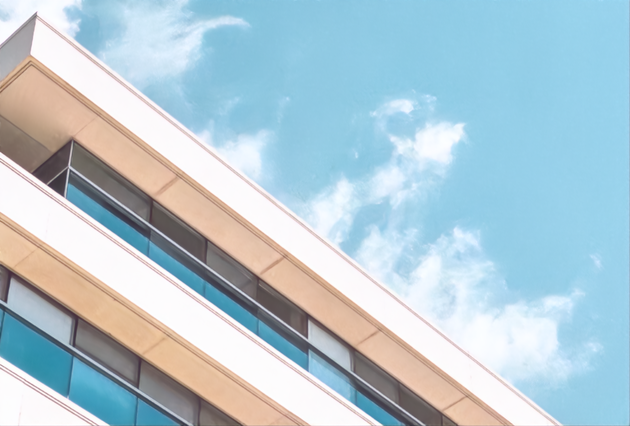}
    \caption*{NAFNet}
\end{subfigure}
\begin{subfigure}{.24\linewidth}
    \centering
    \includegraphics[trim=0cm 0cm 0cm 0cm,clip=true,width=\linewidth]{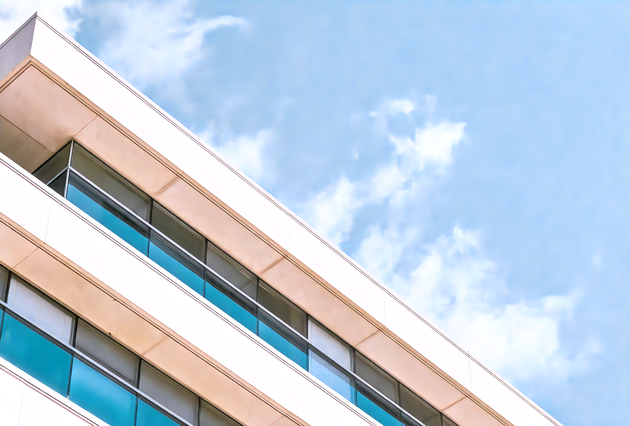}
    \caption*{BSRGAN}
\end{subfigure} 

\vspace{-5px}
\caption*{\footnotesize (b) Image denoising}
\vspace{5px}

\begin{subfigure}{.24\linewidth}
    \centering
    \includegraphics[trim=0cm 0cm 0cm 0cm,clip=true,width=\linewidth]{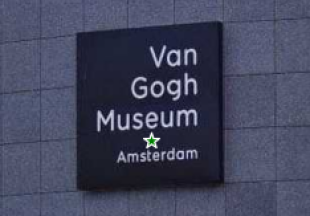}
    \caption*{Input}
\end{subfigure}
\begin{subfigure}{.24\linewidth}
    \centering
    \includegraphics[trim=0cm 0cm 0cm 0cm,clip=true,width=\linewidth]{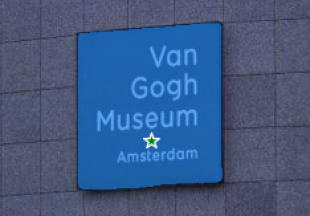}
    \caption*{SAM}
\end{subfigure}
\begin{subfigure}{.24\linewidth}
    \centering
    \includegraphics[trim=0cm 0cm 0cm 0cm,clip=true,width=\linewidth]{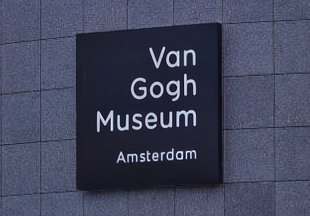}
    \caption*{RAP (Ours)}
\end{subfigure}
\begin{subfigure}{.24\linewidth}
    \centering
    \includegraphics[trim=0cm 0cm 0cm 0cm,clip=true,width=\linewidth]{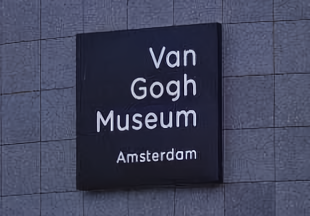}
    \caption*{FBCNN}
\end{subfigure} 

\begin{subfigure}{.24\linewidth}
    \centering
    \includegraphics[trim=0cm 0cm 0cm 0cm,clip=true,width=\linewidth]{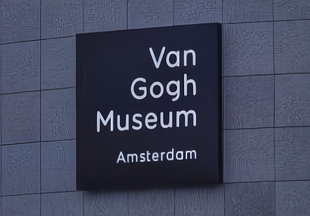}
    \caption*{FBCNN-Strong}
\end{subfigure}
\begin{subfigure}{.24\linewidth}
    \centering
    \includegraphics[trim=0cm 0cm 0cm 0cm,clip=true,width=\linewidth]{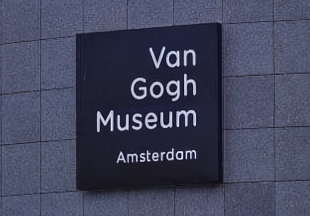}
    \caption*{DNCNN}
\end{subfigure}
\begin{subfigure}{.24\linewidth}
    \centering
    \includegraphics[trim=0cm 0cm 0cm 0cm,clip=true,width=\linewidth]{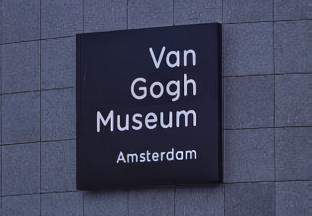}
    \caption*{DPIR}
\end{subfigure}
\begin{subfigure}{.24\linewidth}
    \centering
    \includegraphics[trim=0cm 0cm 0cm 0cm,clip=true,width=\linewidth]{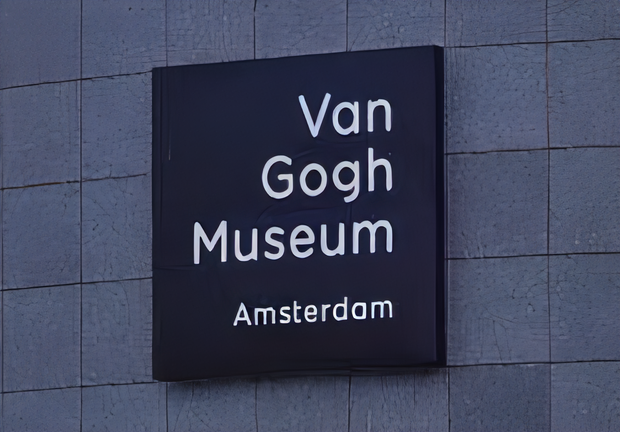}
    \caption*{BSRGAN}
\end{subfigure} 

\vspace{-5px}
\caption*{\footnotesize (c) JPEG artifact removal}
\caption{Visual comparisons of RAP and state-of-the-art methods for image deblurring, image denoising, and JPEG artifact removal. By segmenting the desired object as the first step, our proposed method can apply restoration with different levels to different objects. Please zoom in for visual details.}
\label{fig:comparisons}
\end{figure*}

\section{Experiments}
\subsection{Implementation Details}
We employ DIV2K~\cite{agustsson2017ntire} and Flickr2K~\cite{timofte2017ntire} as our training data.
In the image deblurring task, we used an isotropic Gaussian blur kernel with kernel size $15$ and kernel width $\sigma$ from $0.1$ to $3$. 
For image denoising, we included additive white Gaussian noise with standard deviation from $0$ to $50$.
In the JPEG artifact removal task, we used quality factors from $10$ to $90$.
During training, we randomly extract patch pairs with the size $96\times96$, and the quality factor is randomly sampled from $10$ to $95$. We set $\lambda$ to $0.1$.
To optimize the parameters, we adopt the Adam solver~\cite{kingma2014adam} with batch size 256. 
The learning rate starts from $1 \times 10^{-4}$ and decays by a factor of 0.5 every $4 \times 10^4$ iterations and finally ends with $1.25 \times 10^{-5}$
. 
We trained our model with PyTorch on two NVIDIA GeForce GTX 3090 GPUs.
Training took about two days to finish.

\subsection{Comparisons to State-of-the-Art Methods}
We show the experimental results of RAP on three tasks: image deblurring, image denoising, and JPEG artifact removal.
Figure~\ref{fig:comparisons} illustrates our results, showing that our proposed \emph{Restore Anything Pipeline} can effectively restore low-quality images in an interactive and flexible way.

In the provided image deblurring example, we compare our method with the adapted FBCNN~\cite{jiang2021towards}, Restormer~\cite{zamir2022restormer}, Real-ESRGAN~\cite{wang2021real}, GFPGAN~\cite{wang2021towards}, and BSRGAN~\cite{zhang2021designing}.
From the results, we can observe that Restormer only deblurs with a limited degree. 
On the contrary, the result of Real-ESRGAN is over-smoothed and all textures details are lost. 
GFPGAN presents an overall clear result, but it also changes the appearance of the person, which actually does not restore the image.
BSRGAN provides a result closer to the original one, but the appearance of eyes are also slightly changed.
Instead, FBCNN is only trained with blurry-clear image pairs, so it preserves the most original details. 
Our approach RAP applies FBCNN to the segmented human, making the foreground subject stand out.
To further enhance the image, we apply a blur kernel to the background, thereby producing a bokeh effect.

In the image denoising example, we compare RAP with the state-of-the-art image methods Restormer~\cite{zamir2022restormer}, NAFNet~\cite{chen2022simple}, BSRGAN~\cite{zhang2021designing}, and our adapted FBCNN~\cite{jiang2021towards}.
While we can see that existing methods can remove noise effectively, several side-effects stand out on closer inspection.
Looking at the walls of the building, existing work also removes the texture details and produces over-smoothed results.
Instead, our approach RAP extracts the foreground building and denoises the building and background sky with different restoration levels, thereby removing obvious noise and preserving the high-frequency details at the same time.

To evaluate the performance of RAP on JPEG images, we conducted a comparative analysis with state-of-the-art models, such as the original FBCNN~\cite{jiang2021towards}, DPIR~\cite{zhang2021plug}, and BSRGAN~\cite{zhang2021designing}.
Our observations reveal that DPIR produces remarkable artifact removal, since it is trained on images only with quality factors ranging from 10 to 40.
Despite setting the quality factor to 40, which corresponds to the least restoration level, DPIR tends to overly smooth the wall.
BSRGAN demonstrates a strong restoration capability on the areas near the words.
It can also generate realistic texture details on the wall.
However, these details are synthetically generated and do not accurately represent the original texture. 
On the other hand, the default outputs of FBCNN and DNCNN exhibit a moderate restoration effect.
Notably, FBCNN can be manually adjusted to achieve a higher level of artifact removal, albeit at the expense of losing texture details on the wall. 
Due to RAP's segmentation step, our approach effectively applies a strong restoration level to objects with low-frequency information where artifacts are easily noticeable.
Simultaneously, it employs a weaker restoration level for objects with high-frequency details that require preservation.

\section{Conclusions}
We have presented the \emph{Restore Anything Pipeline} (RAP), an approach to interactive and per-object level image restoration that incorporates image segmentation through SAM into a controllable restoration model. 
Our experiments highlight the adaptability of RAP in addressing image denoising, JPEG artifact removal, and image deblurring tasks.
Our findings reveal that RAP surpasses existing methods in terms of visual results.
This makes RAP a promising avenue for image restoration research, granting users greater control and facilitating the restoration of images at a per-object level.

{\small
\bibliographystyle{ieee_fullname}
\bibliography{main}
}

\end{document}